\documentclass[a4paper, 10pt, conference]{ieeeconf}      

\usepackage{amsmath,amsfonts}
\usepackage{array}
\usepackage[caption=false,font=normalsize,labelfont=sf,textfont=sf]{subfig}
\usepackage{textcomp}
\usepackage{stfloats}
\usepackage{url}
\usepackage{verbatim}
\usepackage{graphicx}
\usepackage{cite}
\usepackage{cleveref} 
\usepackage[ruled,vlined]{algorithm2e}
\usepackage{xcolor}        

\DeclareMathOperator*{\argmin}{arg\,min}

\begin{document}
\IEEEoverridecommandlockouts   
\overrideIEEEmargins           

\title{DELIVER: A System for LLM-Guided Coordinated Multi-Robot Pickup and Delivery using Voronoi-Based Relay Planning}

\author{
    Alkesh K. Srivastava, Jared Michael Levin, Alexander Derrico,
    and Philip Dames  
\thanks{Alkesh Kumar Srivastava, Jared Michael Levin, Alexander Derrico, and Philip Dames are with the Department of Mechanical Engineering, Temple University, Philadelphia, PA, 19122 USA. (e-mail: alkesh@temple.edu; jlevin378@gmail.com; alexander.derrico@temple.edu, pdames@temple.edu).}}


\maketitle
\begin{abstract}
We present DELIVER \textit{(Directed Execution of Language-instructed Item Via Engineered Relay)}, a fully integrated framework for cooperative multi-robot pickup and delivery driven by natural language commands. DELIVER unifies natural language understanding, spatial decomposition, relay planning, and motion execution to enable scalable, collision-free coordination in real-world settings. Given a spoken or written instruction, a lightweight instance of LLaMA3 interprets the command to extract pickup and delivery locations. The environment is partitioned using a Voronoi tessellation to define robot-specific operating regions. Robots then compute optimal relay points along shared boundaries and coordinate handoffs. A finite-state machine governs each robot’s behavior, enabling robust execution. We implement DELIVER on the MultiTRAIL simulation platform and validate it in both ROS2-based Gazebo simulations and real-world hardware using TurtleBot3 robots. {Empirical results show that DELIVER maintains consistent mission cost across varying team sizes while reducing per-agent workload by up to 55\% compared to a single-agent system}. Moreover, the number of active relay agents remains low even as team size increases, demonstrating the system’s scalability and efficient agent utilization. These findings underscore DELIVER’s modular and extensible architecture for language-guided multi-robot coordination, advancing the frontiers of cyber-physical system integration.
\end{abstract}


\section{Introduction}
\begin{figure*}[t]
\centering
\includegraphics[width=\textwidth]{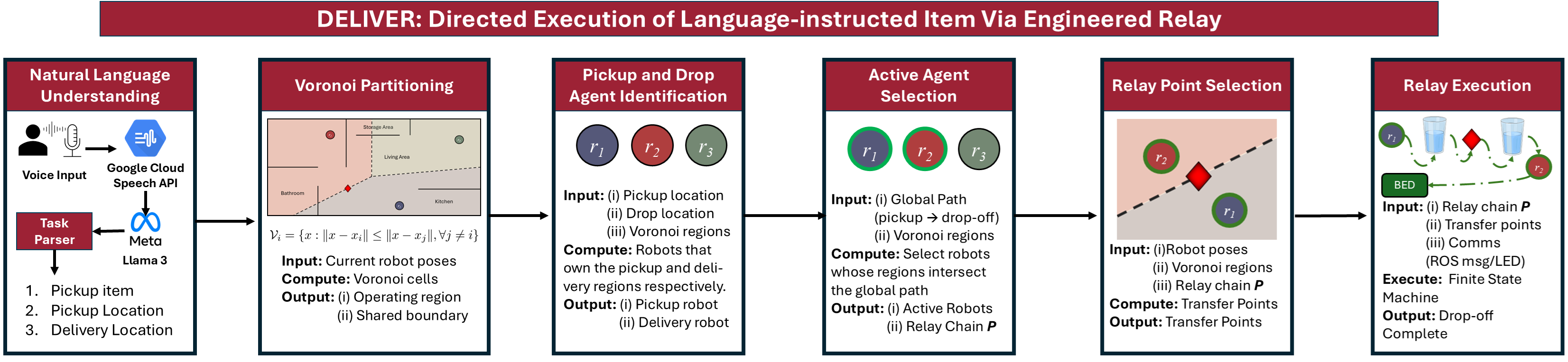}
\caption{System overview of the DELIVER framework. A natural language instruction is processed through speech recognition and LLM-based task parsing to extract pickup and delivery details. The environment is partitioned into Voronoi regions, enabling identification of pickup and drop-off agents. Active agents are then selected based on path-region intersections, and relay points are computed along shared Voronoi boundaries. Finally, robots execute their assigned segments via a finite-state machine, coordinating handoffs through ROS messages or LED cues to complete the delivery.}

\label{fig:concept}
\end{figure*}
Pickup and delivery tasks are fundamental to mobile robotics, forming the core application ranging from warehouse logistics to hospital supply chains~\cite{hossain2023autonomous}.  In large-scale real-world systems, such operations are often distributed, requiring agents to operate within specific regions, coordinate relay handovers, and respond to high-level human tasking. Translating this complexity to autonomous multi-robot teams poses a fundamental challenge: how can a team of robots interpret natural language instructions and collaboratively execute a sequence of handovers?

As robots transition from isolated automation systems to multiple collaborators, the need for intelligent, interpretable, and modular systems integration becomes paramount. In multi-robot environments, enabling teams of robots to collectively execute tasks based on human instructions, without relying on centralized control, is a foundational challenge for next-generation cyber-physical systems.

A common real-world scenario involving pickup and delivery tasks is found in healthcare settings. Mobile robots are increasingly deployed to transport lab samples, deliver medications to patient wards, and relay sterilized equipment between surgical theaters~\cite{holland2021service}. Traditional approaches to these tasks often rely on task allocation, deterministic communication, and rigid, formalized language inputs. However, to be truly autonomous and collaborative, robots must be capable of interpreting natural language, coordinating in a centralized manner, and executing tasks using real-world sensing and control.

We present DELIVER (Directed Execution of Language-Instructed Item Via Engineered Relay), a fully integrated system for cooperative multi-robot pickup and delivery based on natural language commands. As shown in~\Cref{fig:concept}, DELIVER unifies natural language understanding via Meta AI’s LLaMA3 model, Voronoi-based spatial decomposition for region-aware planning, relay-point computation for inter-agent coordination, and execution through local finite-state machines with lightweight signaling. Given a user-issued instruction, DELIVER identifies pickup and delivery locations, plans a global path, segments it across a subset of active agents, executes the relay, and delivers the package to its destination.

This work makes three key contributions: (1) we introduce a fully integrated system that uses language-guided multi-robot architecture combining LLM-based natural language understanding with FSM-based execution; (2) we propose a novel relay coordination mechanism leveraging Voronoi boundaries and modality-flexible signaling (e.g., ROS messages or LED cues) for multi-agent handover; and (3) we validate the system in simulation and hardware using MultiTRAIL,\footnote{A simulation and experimentation platform developed by the Temple Robotics and Artificial Intelligence Lab (TRAIL) at Temple University} ROS2/Gazebo, and TurtleBot3 robots. Experiments show DELIVER reduces per-agent workload by up to 55\% compared to single-agent baselines, while maintaining consistent mission cost and low coordination overhead.

\section{Related Work}

We review prior literature relevant to three core components of our system: 1) Natural Language Understanding (NLU) in Autonomous Navigation, 2) Multi-Agent Pickup and Delivery (MAPD), and 3) Voronoi-based spatial decomposition for multi-agent coordination. Together, these threads inform the design of DELIVER.

\subsection{NLU in Autonomous Navigation}
Recent advances in large language models (LLMs) have transformed natural language understanding (NLU) across domains, including robotics~\cite{zhao2023survey, chang2024survey}. Their ability to generalize and handle unstructured input has proven valuable for collaborative and autonomous systems~\cite{srivastava2024speech, liang2022holistic, nie2019adversarial}. LLMs have achieved state-of-the-art results on tasks such as question answering~\cite{singhal2023towards}, machine translation~\cite{brants2007large}, and text classification~\cite{chae2023large}, with growing applications in task planning and control~\cite{zhou2023comprehensive}.

Several recent systems have explored LLM-driven planning and navigation. SayPlan~\cite{rana2023sayplan} uses 3D scene graphs for scalable task grounding but lacks speech execution. FollowNet and LM-Nav~\cite{shah2018follownet, shah2023lm} apply end-to-end models to connect language to motion in structured settings. GOAT~\cite{chang2023goat} combines object recognition and descriptions, while Arena 3.0~\cite{kastner2024arena} provides simulation tools for social navigation but without advanced NLU integration.

\subsection{Multi-Agent Pickup And Delivery} 

The Multi-Agent Pickup and Delivery (MAPD) problem extends classical routing problems like the Traveling Salesman Problem (TSP) and Multi-Agent Path Finding (MAPF), where agents must repeatedly execute pickup and delivery tasks while avoiding collisions~\cite{bektas2006multiple, croes1958method}. Offline MAPD approaches assign task sequences to agents using variations of mTSP solvers and plan paths accordingly~\cite{liu2019task}. Online variants, such as Token Passing (TP) and Token Passing with Task Swaps (TPTS), support lifelong task streams and decentralized execution~\cite{ma2017lifelong}. Rolling-Horizon Collision Resolution (RHCR) decomposes the problem into windowed subproblems for scalability in dense environments~\cite{li2021lifelong}.

Recent work also considers practical constraints such as agent capacity and multi-task assignment~\cite{kudo2023tsp}, and introduces multi-goal task variants that generalize classical MAPD formulations~\cite{xu2022multi}. These efforts improve throughput and planning efficiency using techniques like Large Neighborhood Search and Priority-Based Search. While most MAPD formulations assume that each task is completed by a single agent, some related domains, such as energy replenishment via rendezvous with mobile chargers~\cite{mathew2015multirobot,li2019rendezvous}, involve multi-agent coordination. However, these approaches are typically limited to energy logistics and do not generalize to relay-based object delivery or execution of user-defined transport tasks, which remain largely unaddressed in MAPD literature.

\subsection{Voronoi-Based Partitioning} 
Voronoi partitioning has been extensively used in multi-robot systems to support a wide range of multi-agent behaviors. Applications such as adaptive sampling~\cite{kemna2017multi}, hazard mapping~\cite{srivastava2022distributed}, GPS-denied coverage~\cite{munir2024anchor}, and collision-aware exploration in unknown terrains~\cite{hu2020voronoi}. Variants of Voronoi diagrams have also been employed for area coverage under connectivity constraints~\cite{luo2019voronoi}, orchard task allocation~\cite{kim2020voronoi}, path-priority planning~\cite{huang2021path}, and non-convex workspace decomposition~\cite{lee2024adaptive,nair2020gm}. Additional efforts extend Voronoi logic to environments with sensor limitations~\cite{guruprasad2012distributed}, target tracking~\cite{dames2020distributed}, and drift-based dynamic costs~\cite{bakolas2010zermelo}. These efforts underscore the versatility of Voronoi methods for scalable, decentralized control across diverse domains. However, to the best of our knowledge, Voronoi-based approaches have not yet been applied to the cooperative multi-agent pickup and delivery (MAPD) problem, leaving a critical gap that this work addresses.

\subsection{Our Contribution}

Building on our previous work on language-guided single-agent navigation~\cite{srivastava2024speech}, DELIVER extends these ideas into a multi-agent context by introducing a relay-based architecture for cooperative pickup and delivery. In contrast to prior work in MAPD and multi-robot coordination, DELIVER introduces a novel relay-based architecture that integrates language understanding, and region-aware planning. The following sections describe the technical design and implementation of the system.

\section{Problem Formulation}
\label{sec:problem_formulation}

We address the problem of language-instructed sequential planning for multi-robot pickup and delivery through a relay-based execution. Given a natural language command from a user, the objective is to coordinate a team of robots to (1) identify the pickup and drop-off locations, (2) assign a sequence of agents to relay the object across the environment, and (3) ensure successful delivery via FSM-control.

Formally, let \( C \) be a user-issued command (e.g., “bring a glass of water from the kitchen to the bedroom”), and let \( \mathcal{R} = \{r_1, \dots, r_n\} \) denote a set of mobile robots operating in a shared workspace \( \mathcal{E} \subset \mathbb{R}^2 \). The system must compute:
\begin{itemize}
     \item A structured task tuple \( (l_{\text{pickup}}, l_{\text{drop}}, I) \in \mathcal{E}\times\mathcal{E} \times \mathcal{I} \), where \( I \) is the item to be transferred;
    \item Voronoi partitions \( \mathcal{V} = \{\mathcal{V}_1, \dots, \mathcal{V}_n\} \) of the environment \( \mathcal{E} \), where each region \( \mathcal{V}_i \) is assigned to robot \( r_i \);
    \item A subset of active relay agents \( \mathcal{P} \subseteq \mathcal{R} \) whose Voronoi regions intersect the planned task path;
    \item A sequence of transfer points \( \{z_1, \dots, z_k\} \subset \mathcal{E} \) along the shared Voronoi boundaries between agents in \( \mathcal{P} \).
\end{itemize}
\begin{figure}[t]
    \centering
    \includegraphics[width=0.48\textwidth]{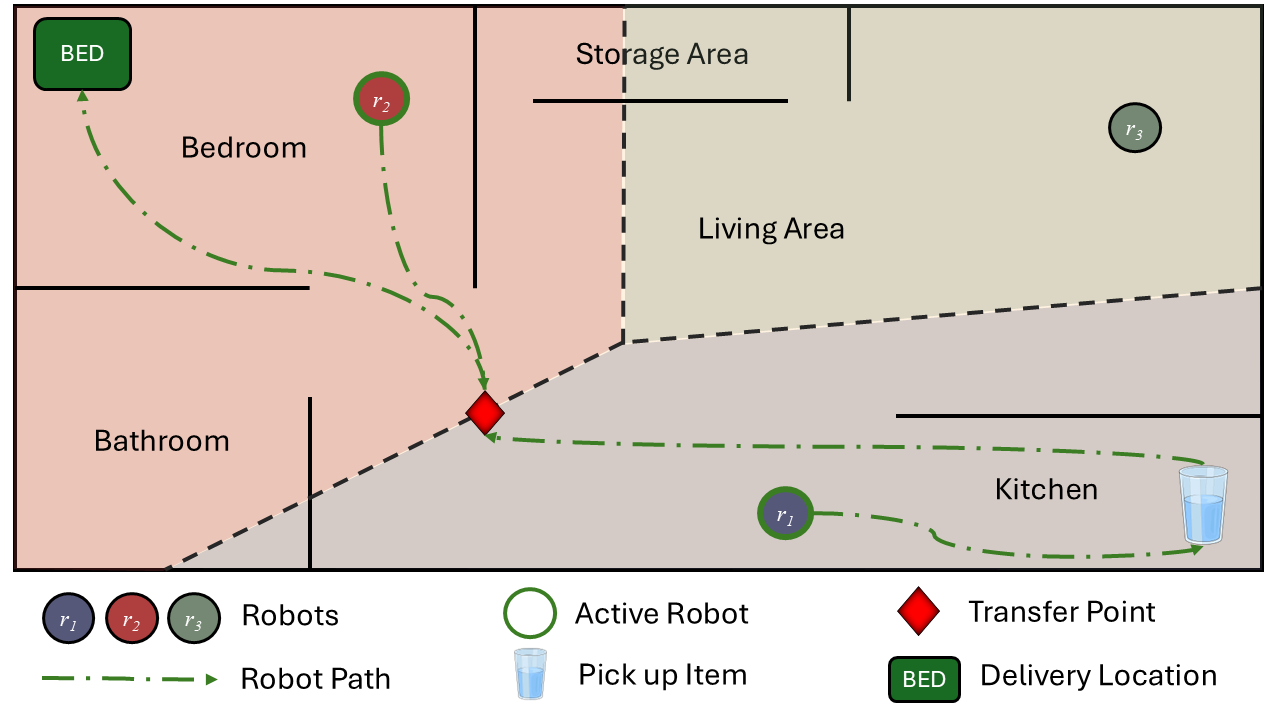}
    \caption{Example of a language-instructed multi-robot relay task. Robot \( r_1 \) picks up an item from the Kitchen and delivers it to a transfer point, where robot \( r_2 \) takes over and completes the delivery to the Bedroom. Non-participating robot \( r_3 \) remains idle.}
    \label{fig:layout_example}
\end{figure}

Each robot \( r_i \in \mathcal{P} \) must execute a local trajectory to its assigned transfer point, coordinate handover via signaling (e.g., LED cues), and proceed through its finite-state controller until delivery is complete.

\paragraph{Illustrative Example}

Consider the scenario presented in~\Cref{fig:layout_example} with a robot tasked with the instruction: \textit{“Bring the glass of water from the kitchen to the bedroom.”} DELIVER parses this natural language command to extract the pickup location (Kitchen), drop-off location (Bedroom), and the object (glass of water). Three robots \( \mathcal{R} = \{r_1, r_2, r_3\} \) are available in the environment, and a Voronoi partition \( \mathcal{V} \) is computed based on their positions.

Based on the location of pickup and delivery, active agents \( \mathcal{P} = \{r_1, r_2\} \) are selected. A transfer point is computed along the shared boundary between their regions. Robot \( r_1 \) navigates to the kitchen, picks up the item, and moves to the transfer point. Upon arrival, it signals readiness using either LED cues or a ROS message. Robot \( r_2 \), upon perceiving this signal, proceeds to complete the relay by transporting the object to the bedroom. This end-to-end coordination process is illustrated in~\Cref{fig:layout_example}, which shows the room layout, robot positions, Voronoi regions, global path, and the active transfer point.

The core challenge lies in integrating natural language understanding, spatial partitioning, relay coordination, and path planning into a scalable cyber-physical system. In the following section, we present the DELIVER architecture, which addresses this problem through six modular components.

\section{DELIVER: Directed Execution of Language-Instructed Item via Engineered Relay}
\label{sec:architecture}

The \textbf{DELIVER} system (\textit{Directed Execution of Language-instructed Item Via Engineered Relay}) addresses the cooperative multi-robot pickup and delivery (MAPD) task through a sequence of six tightly coupled modules. These modules span human interaction, spatial reasoning,  and motion planning, as summarized in~\Cref{fig:concept}. Together, they form a robust cyber-physical system that enables language-guided coordination through relay-based planning. We formulate each component of the system as a distinct subproblem and describe the corresponding solution integrated within DELIVER.

\subsection{Natural Language Understanding via LLM}
\noindent\textbf{Objective:} Translate a natural language command \( C \) into a structured task representation:
\[
(l_{\text{pickup}}, l_{\text{drop}}, I) = \texttt{LLM}(C)
\]
where \( l_{\text{pickup}}, l_{\text{drop}} \in \mathbb{R}^2 \) are pickup and drop-off coordinates, and \( I \) is the item to be delivered.

\noindent\textbf{Solution:}  
We use Meta AI’s LLaMA3-8B model, queried with user instructions transcribed via the Google Speech API. The output is parsed using regex-based heuristics to extract semantic fields, avoiding rigid templates. This module is directly adopted from our prior work~\cite{srivastava2024speech}, where we observed an average task classification accuracy of \textbf{84.37\%} across diverse user-spoken commands. These results validate the robustness of our LLM-based parser and motivate its use in the current multi-robot setting to bridge high-level natural language input with spatial task planning.
\subsection{Voronoi-Based Spatial Partitioning}
\noindent\textbf{Objective:} Partition the workspace among robots to define geometric responsibility:
\[
\mathcal{V}_i = \left\{ x \in \mathcal{E} \mid \|x - x_i\| \leq \|x - x_j\|, \forall j \neq i \right\}
\]

\noindent\textbf{Solution:}  
In typical warehouse, office, or home-like indoor environments where obstacles are sparse and relatively small, we approximate the workspace as obstacle-free and convex for high-level spatial allocation. This simplification enables efficient Voronoi-based partitioning, while any deviations caused by actual obstacles are handled by low-level planning or control (e.g., ROS 2 Nav2). We generate a Voronoi diagram over the robot positions to partition the environment into spatial regions \( \mathcal{V}_i \), where each region is assigned to a robot \( r_i \).  These regions define spatial ownership and serve as the basis for downstream relay assignment and region-aware coordination. While our current formulation assumes convex environments for simplicity, we expect that in obstacle-rich environments, global path planning or low-level controllers (e.g., within MultiTRAIL or ROS 2) will handle obstacle avoidance, allowing the high-level Voronoi-based coordination scheme to remain valid.

\subsection{Pickup and Drop Agent Identification}
\noindent\textbf{Objective:} Determine the initiating and terminating robots responsible for the pickup and drop-off locations.



\noindent\textbf{Solution:}  
We use the Voronoi partition \( \mathcal{V} = \{\mathcal{V}_1, \dots, \mathcal{V}_n\} \) of the environment to assign responsibility for the task endpoints. Given a pickup location \( l_{\text{pickup}} \) and a drop-off location \( l_{\text{drop}} \), the corresponding robots are:
\[
r_{\text{pickup}} = r_i \text{ such that } l_{\text{pickup}} \in \mathcal{V}_i, \]
\[
r_{\text{drop}} = r_j \text{ such that } l_{\text{drop}} \in \mathcal{V}_j
\]
This ensures that each endpoint is assigned to the robot whose Voronoi region contains it. In edge cases where a location lies on a boundary, proximity-based tie-breaking (minimal Euclidean distance) is used.

\subsection{Active Agent Selection}\label{subsec:active_agents}
\noindent\textbf{Objective:} Determine which robots are responsible for handling the relay by checking path-region intersections.

Let \( \pi \) denote the A*-planned path from \( l_{\text{pickup}} \) to \( l_{\text{drop}} \). The set of active agents is:
\[
\mathcal{P} = \{r_i \in \mathcal{R} \mid \pi \cap \mathcal{V}_i \neq \emptyset \}
\]

\noindent\textbf{Solution:}  
We run A* planning over the entire environment to compute a global path \( \pi \) from the pickup to the drop-off location. This path is then overlaid on the Voronoi partition \( \mathcal{V} = \{\mathcal{V}_i\} \), and the selection of active agents is based on which regions intersect with \( \pi \):
\[
\mathcal{P} = \{r_i \in \mathcal{R} \mid \pi \cap \mathcal{V}_i \neq \emptyset \}
\]
Robots whose regions intersect the path are identified as active agents and assigned to relay segments of the task. This strategy localizes coordination to a minimal subset of the team. An example is shown in~\Cref{fig:layout_example}, where robots \( r_1 \) and \( r_2 \) are identified as active agents for relaying the glass of water from the kitchen to the bedroom.
\subsection{Relay Transfer Point Computation}
\noindent\textbf{Objective:} For each pair \( (r_i, r_{i+1}) \in \mathcal{P} \), compute
\[
z^*=\argmin_{z\in\Gamma_{i,i+1}} \max\!\big(\|z-x_i\|,\|z-x_{i+1}\|\big),
\]
where \(\Gamma_{i,i+1}\) is the shared Voronoi boundary (edge segment).

\noindent\textbf{Solution:} The optimal point is the intersection of the perpendicular bisector of \(x_i\) and \(x_{i+1}\) with \(\Gamma_{i,i+1}\). Let endpoints \(p_1,p_2\) define the edge, \(e=p_2-p_1\), midpoint \(m=\tfrac{1}{2}(x_i+x_{i+1})\), difference \(d=x_{i+1}-x_i\), and bisector direction \(b=\begin{bmatrix}-d_y & d_x\end{bmatrix}^T\). 

\noindent Solve
\(
p_1+t\,e = m+s\,b
\quad\Longleftrightarrow\quad
\begin{bmatrix} e & -b \end{bmatrix}\!\begin{bmatrix} t\\ s \end{bmatrix}=m-p_1.
\)
If \(0\le t\le 1\), set \(z^*=p_1+t\,e\); otherwise choose
\[
z^*\in\argmin_{p\in\{p_1,p_2\}} \max\!\big(\|p-x_i\|,\|p-x_{i+1}\|\big).
\]





\subsection{Relay Execution}
\noindent\textbf{Objective:} Execute each robot's segment and perform handovers at transfer points using lightweight signaling.

\noindent\textbf{Solution:}  
Each robot plans its local trajectory based on its position in the relay chain:
\begin{itemize}
    \item \textbf{Initiator (first robot):} \( x_i \rightarrow l_{\text{pickup}} \rightarrow z_1 \)
    \item \textbf{Relay robots (intermediate agents):} \( x_i \rightarrow z_{j} \rightarrow z_{j+1} \)
    \item \textbf{Final robot:} \( x_i \rightarrow z_k \rightarrow l_{\text{drop}} \)
\end{itemize}
Here, \( z_j \) and \( z_{j+1} \) denote consecutive transfer points in the relay chain.

At each transfer point, the sending robot signals readiness via a ROS message or LED color change. The receiving robot, upon detecting this signal, initiates its motion. Each robot operates a local finite-state machine (FSM) with states: \texttt{Idle}, \texttt{Navigate}, \texttt{Pickup}, \texttt{Relay}, and \texttt{Deliver}. State transitions are triggered by reaching waypoints and observing signaling cues.

\section{Experiments and Results}
\begin{figure*}[t]
    \centering
    \includegraphics[width=\textwidth]{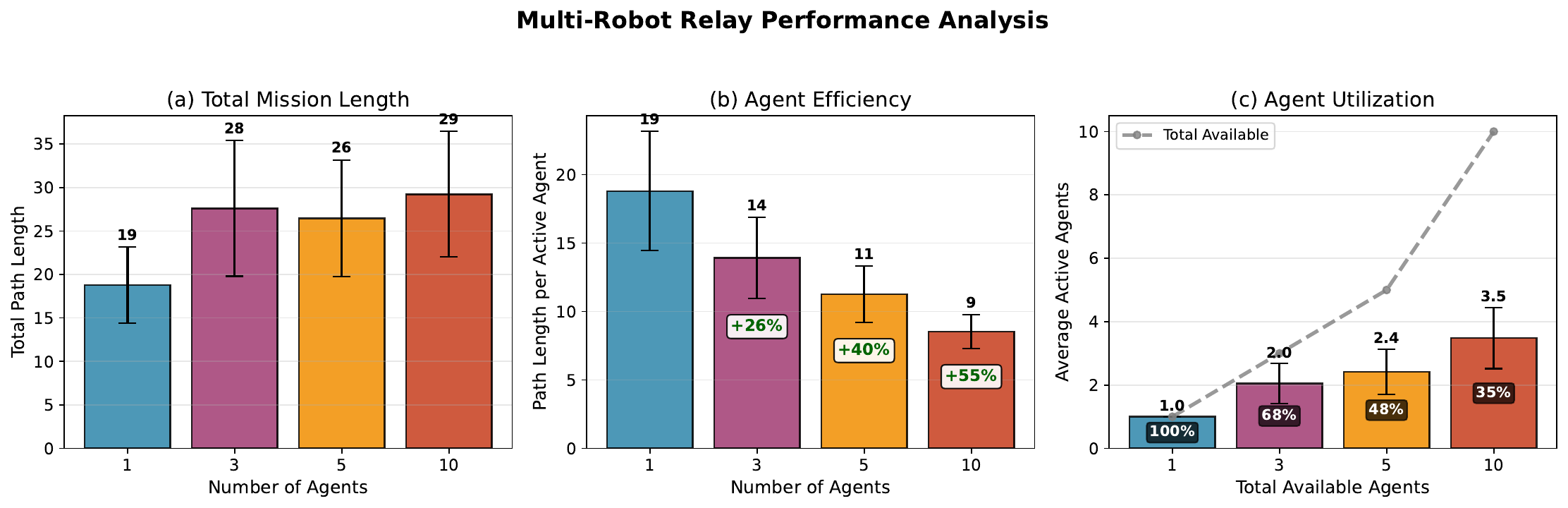}
    \caption{Multi-robot relay performance analysis across varying team sizes ($n = 1, 3, 5, 10$). 
    (a)~Total path length (sum over segments) stays within 19–29 units, indicating stable overall cost. 
    (b)~Path length per active agent decreases with larger teams, demonstrating improved efficiency through workload distribution. 
    (c)~Average number of active relay agents increases sublinearly with team size, indicating efficient agent utilization even as more robots become available.}
    \label{fig:relay_performance}
\end{figure*}
To validate the DELIVER system, we evaluate its performance across three distinct environments: discrete-grid simulation in MultiTRAIL, continuous-space simulation using ROS2 and Gazebo, and real-world deployment on TurtleBot3 robots. These experiments aim to assess system robustness, scalability, and the seamless integration of language understanding, planning, and delivery execution.

\subsection{MultiTRAIL Simulation}

\noindent
\textbf{MultiTRAIL} is a discrete-grid simulation and visualization framework developed by the TRAIL Lab at Temple University to support rapid prototyping and evaluation of multi-robot behaviors. In this environment, we deploy and test the full DELIVER pipeline on a $20 \times 20$ grid with teams of 3, 5, 7, and 10 agents.

For each team size, we run 100 independent trials. In every trial, robots are placed at random non-overlapping start locations, and pickup and delivery coordinates are randomly selected with a minimum separation of eight cells. Each trial executes a single pickup-and-delivery task, initiated in deployment via natural language (parsed by an LLM) but sampled directly in simulation to focus on planning and coordination.  We compare DELIVER against a baseline in which a single robot performs the entire task without relays.

Key evaluation metrics are total path length (sum of segment lengths), number of active relay agents, and mean path length per agent. As shown in~\Cref{fig:relay_performance}, total mission cost remains nearly constant across team sizes, while mean per-agent path decreases by up to 55\%, indicating effective workload distribution. The number of active relay agents grows sublinearly as team size increases, demonstrating scalable coordination without engaging \textit{unnecessary agents}.
\begin{figure}[tbph]
\centering
\includegraphics[width=0.41\textwidth]{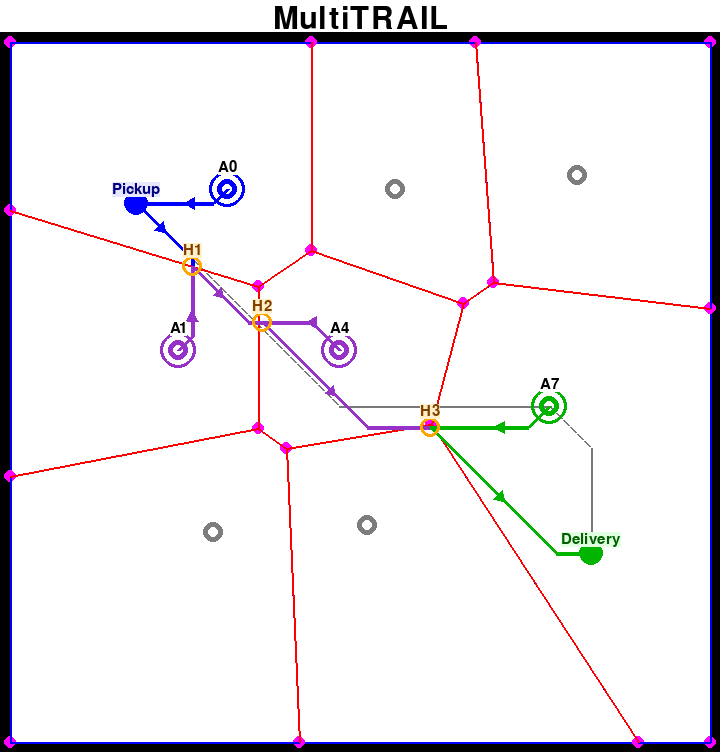}
\caption{Snapshot of the MultiTRAIL simulation environment with 8 agents and their Voronoi-based spatial decomposition (red edges). The pickup agent (A0) is highlighted in blue, with its path to the first handoff (H1) shown in blue. Intermediate relay agents A1 and A4 are active in the relay chain and are depicted in purple, with their corresponding path segments also in purple. The delivery agent (A7) is shown in green, along with its path to the delivery location. The three handoff points, H1–H3, are marked in orange, illustrating the sequential transfer of the package from the pickup agent through successive relays to the final delivery agent.}

\label{fig:multitrail}
\end{figure}

\begin{figure}[tbph]
\centering
\includegraphics[width=\linewidth]{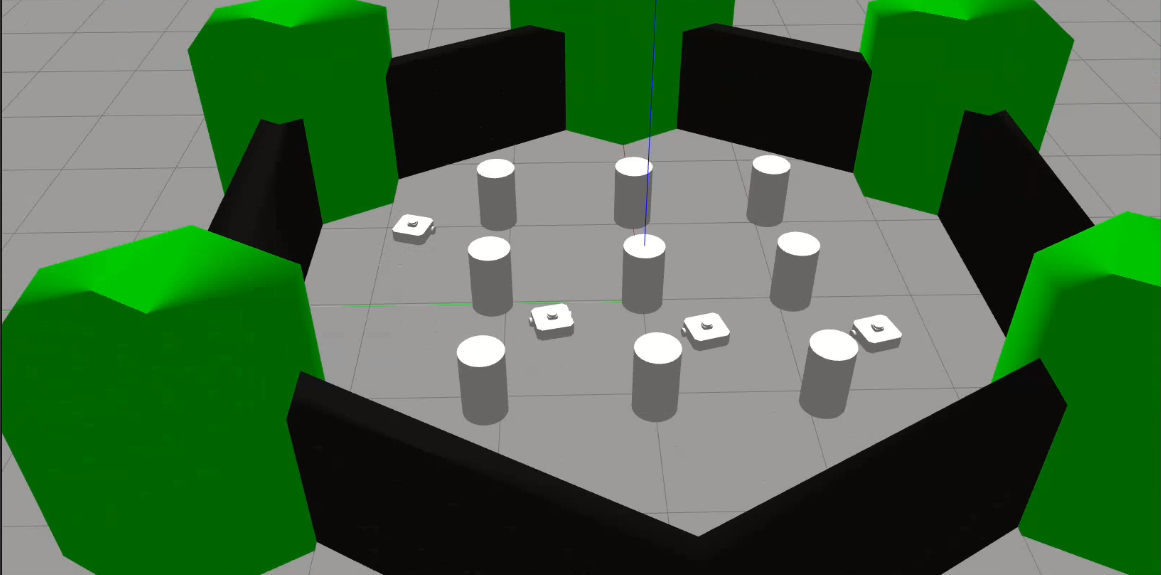}
\caption{ROS2/Gazebo simulation of the DELIVER system with four TurtleBot3 robots navigating a cluttered environment. Robots (white) coordinate to relay a package or information through cylindrical obstacles and narrow passages. Voronoi partitions and relay paths are computed offline in MultiTRAIL based on robot positions in a convex 2D workspace. These paths are provided as waypoints to the ROS2 Nav2 stack, which handles obstacle-aware motion planning and local control. This setup is used to evaluate planning, relay execution, and multi-robot coordination under realistic spatial constraints.}
\label{fig:gazebo}
\end{figure}
\begin{figure}[!t]
    \centering
    \includegraphics[width=0.5\textwidth]{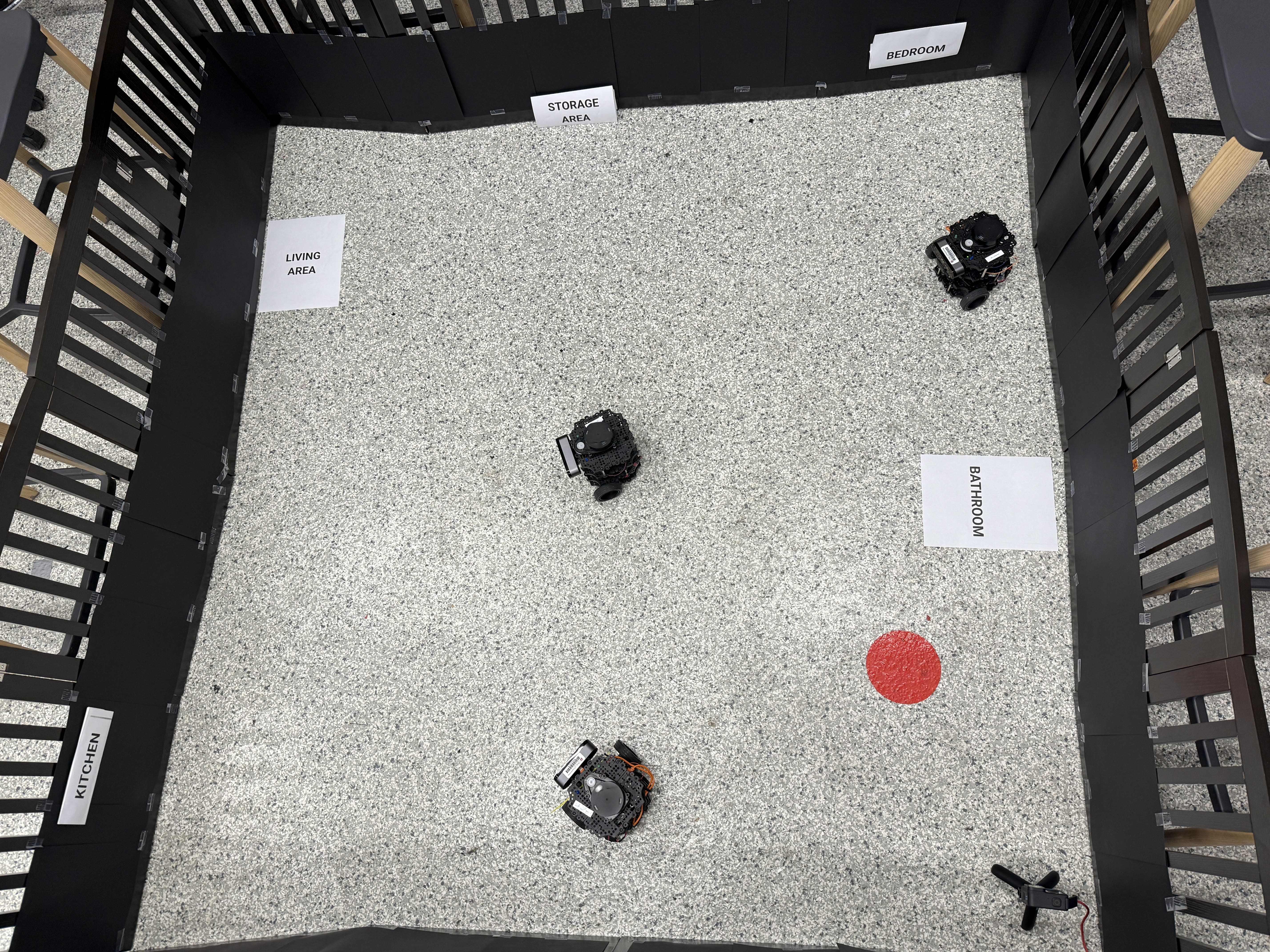}
    \caption{Real-world testbed for \textsc{DELIVER} in a \(1.7 \times 1.7\)\,m enclosed indoor environment. The space is semantically labeled to match natural-language task descriptions.}
    \label{fig:cage}
\end{figure}

\subsection{ROS2/Gazebo Simulation}

To evaluate DELIVER in a continuous-space setting, we simulate the system in Gazebo using the ROS2 Nav2 stack. The environment is the default TurtleBot3 world, which resembles an open warehouse-like space with a mostly unobstructed floor plan and a small number of pillar-like obstacles. While this environment is technically non-convex, for high-level coordination we approximate it as obstacle-free and convex, allowing us to apply the same Voronoi-based spatial allocation used in the discrete setting. Local obstacle avoidance and path feasibility are handled entirely by the Nav2 planners, ensuring that the convex approximation does not affect execution. 

We conducted experiments with 2 and 4 TurtleBot3 robots. Each robot used AMCL for localization and the Smac hybrid-A* planner from Nav2 for obstacle-aware trajectory generation. Robots were assigned pickup and drop-off points based on natural language commands (text-based), and relay handoff locations were determined via Voronoi-based planning.

Each robot executed a local segment of the delivery using a finite-state machine, coordinating handovers via ROS messages. We performed 10 trials for each configuration, and each trial consisted of a single relay-based delivery task.

\subsubsection*{Results}
All trials completed the relay task. The generated paths respected obstacle avoidance and velocity constraints, and FSM transitions occurred reliably at transfer points. While Voronoi partitioning assumes an obstacle-free environment, we observed that the underlying ROS2 stack (including AMCL localization and Nav2 planning) enabled reliable navigation even in cluttered, obstacle-rich environments. The relay coordination and handovers remained consistent across trials. 
\subsection{Real-World Hardware Deployment}
\begin{figure}[!t]
    \centering
    \subfloat[]{\includegraphics[width=0.45\linewidth]{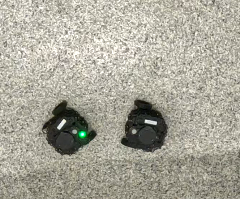}}
    \hfill
    \subfloat[]{\includegraphics[width=0.45\linewidth]{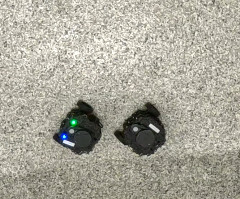}}\\
    \subfloat[]{\includegraphics[width=0.45\linewidth]{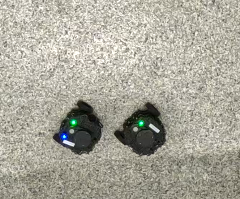}}
    \hfill
    \subfloat[]{\includegraphics[width=0.45\linewidth]{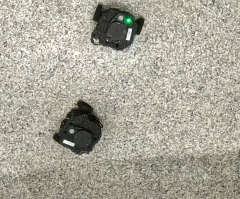}}
    \caption{Example relay handoff sequence in the two-robot configuration. 
    (a) The first robot arrives at the transfer point with the green LED active. 
    (b--c) The receiving robot arrives, and the first robot’s LED switches to blue while the second robot’s LED turns green. 
    (d) The second robot departs toward the next waypoint with the package. Once a robot completes its handoff, it turns off all LEDs to indicate task completion.}
    \label{fig:sequence}
\end{figure}
We validate DELIVER on three physical TurtleBot3 robots operating in a \(1.7 \times 1.7\)\,m enclosed, obstacle-free indoor environment (Fig.~\ref{fig:cage}). The floor space is divided into semantically labeled zones--\textit{Kitchen}, \textit{Living Area}, \textit{Storage Area}, \textit{Bedroom}, and \textit{Bathroom}--to enable natural-language autonomous navigation. This setup is similar to the layout in~\Cref{fig:layout_example}.

In this experiment, LEDs are used solely for visual status confirmation by human observers: green indicates that a robot is currently carrying the package, while blue indicates that the robot has reached the transfer point. When the handoff is successful, both the LEDs are turned off. Relay cueing is handled via JSON-based messages broadcast over Wi-Fi from the central coordinator, which also synchronizes LED state changes across all robots. 

We evaluate two configurations:
\begin{enumerate}
    \item \textbf{Three-Robot Relay:} All three robots participate in the delivery, performing two handoffs between pickup and final drop-off.
    \item \textbf{Two-Robot Relay:} Following the active-agent selection logic described in Sec.~\ref{subsec:active_agents}, only two robots are chosen to execute the delivery while the third remains idle, reducing unnecessary movement.
\end{enumerate}

In both configurations, the MultiTRAIL framework generates the complete relay plan, using the LLM-parsed pickup and drop-off locations together with Voronoi-based relay assignment, and computes the corresponding transfer points. This plan is then passed to the ROS 2 stack, which dispatches waypoints to each active robot for execution. \Cref{fig:sequence} illustrates an example handoff sequence in the two-robot configuration. The robot carrying the package is indicated by its green LED being on. Upon reaching the transfer point, the same robot also turns on its blue LED to signal arrival for handoff. After the package is transferred, both LEDs on the sending robot are turned off, while the receiving robot turns on its green LED to indicate possession. All LED changes are triggered in real time by JSON messages from the central coordinator, ensuring consistent and reliable signaling.

\subsubsection*{Results}
Across all trials in both configurations, tasks were completed successfully without human intervention. JSON-triggered finite-state transitions occurred reliably at the appropriate waypoints, and LED status updates were consistently synchronized under standard laboratory lighting. The observed behavior closely matched the simulation results, confirming that the ROS~2-based control and coordination stack transfers effectively to physical robots. A video demonstration of the experiments is available online.\footnote{\url{https://youtu.be/-IPeJjbi4RA}}




\section{Conclusion}

We present \textbf{DELIVER}, a fully integrated system for cooperative multi-robot pickup and delivery based on natural language instructions and relay coordination. Combining large language models for command interpretation, Voronoi-based spatial reasoning for region segmentation, and a relay handoff mechanism, DELIVER enables scalable collaboration across both simulated and physical platforms. Through extensive evaluation in MultiTRAIL, ROS2/Gazebo simulations, and real-world TurtleBot3 deployments, we demonstrated that DELIVER achieves up to 55\% reduction in per-agent workload while maintaining low coordination overhead. These results highlight the system’s robustness, scalability, and efficient agent utilization. DELIVER exemplifies the core goals of system integration research by tightly coupling perception, planning, control, and human interaction into a unified, extensible task-executing architecture. Future work will focus on dynamic task reassignment, real-time request handling, and operation in human-shared environments to further enhance autonomy, versatility, and real-world viability. In addition, we plan to extend the Voronoi-based spatial allocation to handle non-convex environments with obstacles, addressing edge cases where the optimal transfer point lies within or near an obstacle, thereby improving robustness in more complex workspaces.

\section*{Acknowledgment}

This work was funded by NSF grant CNS-2143312. The authors would like to thank Kevin Rafael Formento and Malin Aaron Kussi for their invaluable assistance with the hardware experiments and TurtleBot3 setup. Their support was instrumental in the successful deployment and validation of the DELIVER system on physical robots.

\bibliographystyle{IEEEtran}
\bibliography{icsr,mapd, voronoi}

\end{document}